\documentclass[pdflatex,sn-mathphys-num]{sn-jnl}
\usepackage{lmodern}
\usepackage[T1]{fontenc}
\usepackage{microtype}
\usepackage{graphicx}
\usepackage{amsmath,amssymb}
\usepackage{booktabs}
\usepackage{placeins}
\usepackage{caption}
\AtBeginDocument{\setlength{\parindent}{0pt}\setlength{\parskip}{5pt plus 1pt minus 1pt}}
\usepackage{etoolbox}
\makeatletter
\patchcmd{\@maketitle}{Corresponding author(s). E-mail(s):}{Corresponding authors:}{}{\PackageWarning{main}{author-email patch failed}}
\renewcommand\normalsize{%
  \@setfontsize\normalsize{11bp}{13.5bp}%
  \abovedisplayskip 12\p@ \@plus2\p@ \@minus1\p@
  \abovedisplayshortskip \z@ \@plus3\p@
  \belowdisplayshortskip 3\p@ \@plus3\p@ \@minus3\p@
  \belowdisplayskip \abovedisplayskip
  \let\@listi\@listI}
\renewcommand\small{\@setfontsize\small{10bp}{12bp}}
\def\abstractheadfont{\reset@font\fontsize{11bp}{13.5bp}\bfseries\selectfont\titraggedcenter}
\def\abstractfont{\reset@font\fontsize{11bp}{13.5bp}\selectfont\leftskip=24pt\rightskip=24pt\parfillskip=0pt plus 1fil}
\def\keywordfont{\reset@font\fontsize{9bp}{11bp}\selectfont\leftskip=24pt\rightskip=24pt plus0.5fill}
\def\bmheadfont{\reset@font\fontfamily{\rmdefault}\fontsize{11bp}{13.5bp}\bfseries\selectfont\raggedright\boldmath}
\def\bibfont{\reset@font\fontfamily{\rmdefault}\small\selectfont}
\normalsize
\patchcmd{\@maketitle}{\else\vskip21pt\fi}{\else\fi}{}{\PackageWarning{main}{title-space patch 1 failed}}
\patchcmd{\@maketitle}{\removelastskip\vskip20pt\nointerlineskip}{\removelastskip\nointerlineskip}{}{\PackageWarning{main}{title-space patch 2 failed}}
\renewcommand\section{\@startsection{section}{1}{\z@}{-24pt \@plus -4pt \@minus -2pt}{6pt}{\sectionfont}}
\renewcommand\subsection{\@startsection{subsection}{2}{\z@}{-16pt \@plus -3pt \@minus -2pt}{4pt}{\subsectionfont}}
\renewcommand\subsubsection{\@startsection{subsubsection}{3}{\z@}{-12pt \@plus -2pt \@minus -2pt}{3pt}{\subsubsectionfont}}
\providecommand{\burl}[1]{\url{#1}}
\def\email#1{\global\advance\emailcnt by 1\relax
  \if@corauemail
    \ifx\corrauthemail\@empty\else\g@addto@macro\corrauthemail{; }\fi
    \g@addto@macro\corrauthemail{\textcolor{blue}{#1}}%
  \else
    \ifx\authemail\@empty\else\g@addto@macro\authemail{; }\fi
    \g@addto@macro\authemail{\textcolor{blue}{#1}}%
  \fi}
\makeatother
\usepackage{fancyhdr}
\usepackage{geometry}
\hypersetup{
  pdftitle={Neurosymbolic disaster assessment from satellite imagery},
  pdfauthor={Arushi Gupta, Tianyuan Huang, Zhecheng Wang, Joy Hsu, Jiajun Wu},
}

\begin{document}

\title[Neurosymbolic disaster assessment from satellite imagery]{\textbf{Neurosymbolic disaster assessment\hb from satellite imagery}}

\author*[1]{\fnm{Arushi} \sur{Gupta}}\email{arugupta@stanford.edu}
\author[1]{\fnm{Tianyuan} \sur{Huang}}
\author[1]{\fnm{Zhecheng} \sur{Wang}}
\author*[1]{\fnm{Joy} \sur{Hsu}}\email{joycj@stanford.edu}\equalcont{These authors jointly supervised this work.}
\author[1]{\fnm{Jiajun} \sur{Wu}}\equalcont{These authors jointly supervised this work.}

\affil[1]{\orgname{Stanford University}, \orgaddress{\city{Stanford}, \state{CA}, \country{USA}}}

\abstract{After disasters, responders use satellite imagery to quickly assess damage and guide response at scale.
As priorities evolve, they need timely answers about affected areas, such as how many homes were destroyed or the density of blocked roads.
Yet answering these questions accurately and efficiently remains challenging.
Here, we introduce SIREN, a neurosymbolic framework that predicts properties of entities such as buildings and roads from satellite imagery and combines these predictions through explicit reasoning to answer diverse questions.
Across 15 disasters spanning six types, SIREN estimates counts more accurately than state-of-the-art remote-sensing foundation models and frontier vision-language models, reducing error by 7.7--28.9\% on questions that combine multiple conditions.
By reusing intermediate results across questions, SIREN reduces model computation by 4.4--27.7-fold compared with baselines.
SIREN also learns effectively from limited annotations and transfers to three disasters without event-specific calibration.
Ultimately, SIREN provides an accurate, efficient and flexible approach to disaster assessment.}

\begingroup
\thispagestyle{empty}
\setlength{\parskip}{0pt}
\maketitle
\endgroup
\makeatletter\global\@nobreakfalse\makeatother

Modern Earth observation systems collect vast amounts of high-resolution satellite imagery before and after natural disasters such as floods, earthquakes and hurricanes \cite{gupta2019xbd,wang2025disasterm3}.
These images provide a detailed view of affected regions and can support disaster response and recovery efforts \cite{internationalcharter2026,copernicusems2026}.
A central challenge is transforming these images into actionable information quickly and at scale.
In practice, responders need answers to queries such as: \emph{How many damaged buildings have no nearby clear road? How many residential buildings were destroyed? How many blocked road segments lie near destroyed buildings?} Answering these queries requires reasoning over entities, their properties and the relationships between them \cite{lobry2020rsvqa,wang2024earthvqa,wang2025disasterm3}.

Existing approaches for satellite-based disaster analysis include task-specific models and foundation models.
Task-specific models include building damage detectors and road segmentation systems \cite{gupta2020rescuenet,weber2020damage,vanetten2019cresi,hansch2022spacenet8}.
Their main limitation is that they are designed for predefined tasks.
As a result, answering more complex queries may require combining predictions from multiple models, while new types of queries may require training additional models.
Foundation models offer two ways to answer disaster-response queries.
First, remote-sensing foundation models, such as SatMAE, SkySense and RemoteCLIP, are pretrained on large collections of satellite and aerial imagery and can be fine-tuned for individual queries \cite{cong2022satmae,reed2023scalemae,guo2023skysense,liu2024remoteclip,wu2025skysenseplus}.
These models can be efficient for a given query, but adapting them to new queries requires additional training.
Second, generative vision-language models, such as GPT and Claude, along with those fine-tuned on remote-sensing data \cite{kuckreja2023geochat,soni2025earthdial}, answer natural-language queries directly from imagery without query-specific training.
However, these models can struggle with the explicit, multi-step reasoning needed to answer complex queries reliably \cite{wang2025disasterm3}.
Processing the same imagery separately for different queries can also be computationally expensive.

Neurosymbolic approaches can address these limitations by separating visual recognition from reasoning.
Neural models identify entities and predict their properties, and symbolic programs then specify and execute the operations needed to answer a query, such as selecting entities that meet a condition and counting them \cite{yi2018nsvqa,mao2019nscl,amizadeh2020reasoning,hsu2023ns3d}.
More recent neurosymbolic methods use language models to generate programs from natural-language queries, and these programs can call visual models and other computational tools \cite{gupta2023visual,suris2023vipergpt,hsu2023s}.
However, much of this work has focused on synthetic scenes, natural photographs and 3D environments.
Disaster assessment presents additional challenges for visual reasoning.
First, entities can be difficult to identify and delineate; for example, buildings are often densely packed, and roads form extended, interconnected networks.
Second, different properties require different forms of visual understanding; for example, residential use must be inferred from a building's appearance and surroundings, whereas damage must be assessed by comparing pre- and post-disaster images.

To address these challenges, we introduce SIREN (Satellite Image REasoning for Natural Disasters), a neurosymbolic framework for accurate and efficient disaster-response analysis.
SIREN consists of three steps (Fig.~\ref{fig:siren_overview}a).
First, it detects individual entities, such as buildings and roads, from a pre-disaster image and represents each entity using its geometry and features from the pre- and post-disaster images.
Second, it translates a user query into a program specifying the conditions to check and the calculations to perform.
Third, it executes the program using learned predictions and geometric calculations.
Learned classifiers predict visual properties, such as building damage, road blockage and residential use (neural execution).
Geometric properties and relationships, such as building size and proximity, are computed directly in code and combined with these predictions to calculate the answer (symbolic execution).
For example, to count damaged buildings with no nearby clear road, the program selects damaged buildings, checks for nearby clear roads and counts those without one.

Across 15 disaster events spanning six disaster types, SIREN achieves lower error than four state-of-the-art remote-sensing foundation models and two frontier vision-language models on a variety of simple and complex queries.
Beyond accuracy, SIREN enables flexible and interpretable disaster analysis by representing entities, their visual properties and relationships explicitly.
This representation allows new queries to be answered by combining existing predictions and symbolic operations, without training a separate model for each query.
Each answer can also be traced to the entities selected and the calculations performed.

SIREN also addresses practical challenges in disaster response: large areas must be analyzed quickly, information needs change as a response progresses, and labeled imagery may not be available for a new disaster.
By computing shared visual predictions once and reusing them across queries, SIREN reduces the computational cost of answering multiple queries over the same imagery.
It also learns effectively from limited annotations and transfers to three real-world disasters outside its training data without event-specific training or calibration.

Together, these results establish SIREN as a framework for accurate, efficient and flexible disaster assessment from satellite imagery.

\section{Results}

\subsection{SIREN accurately recovers building and road damage states across six disaster types}

We first demonstrate that SIREN can recover basic disaster states from satellite imagery.
We focus on three queries central to early disaster response: which buildings are damaged, which buildings are destroyed and which road segments are blocked.
We use held-out pre- and post-disaster image pairs from 15 events in DisasterM3~\cite{wang2025disasterm3}, spanning floods, hurricanes, wildfires, tornadoes, earthquakes and tsunamis.
We report Count-MAE, the average absolute error in the number of qualifying entities per image pair.

\begin{figure}[p]
\centering
\includegraphics[width=\textwidth,height=0.86\textheight,keepaspectratio]{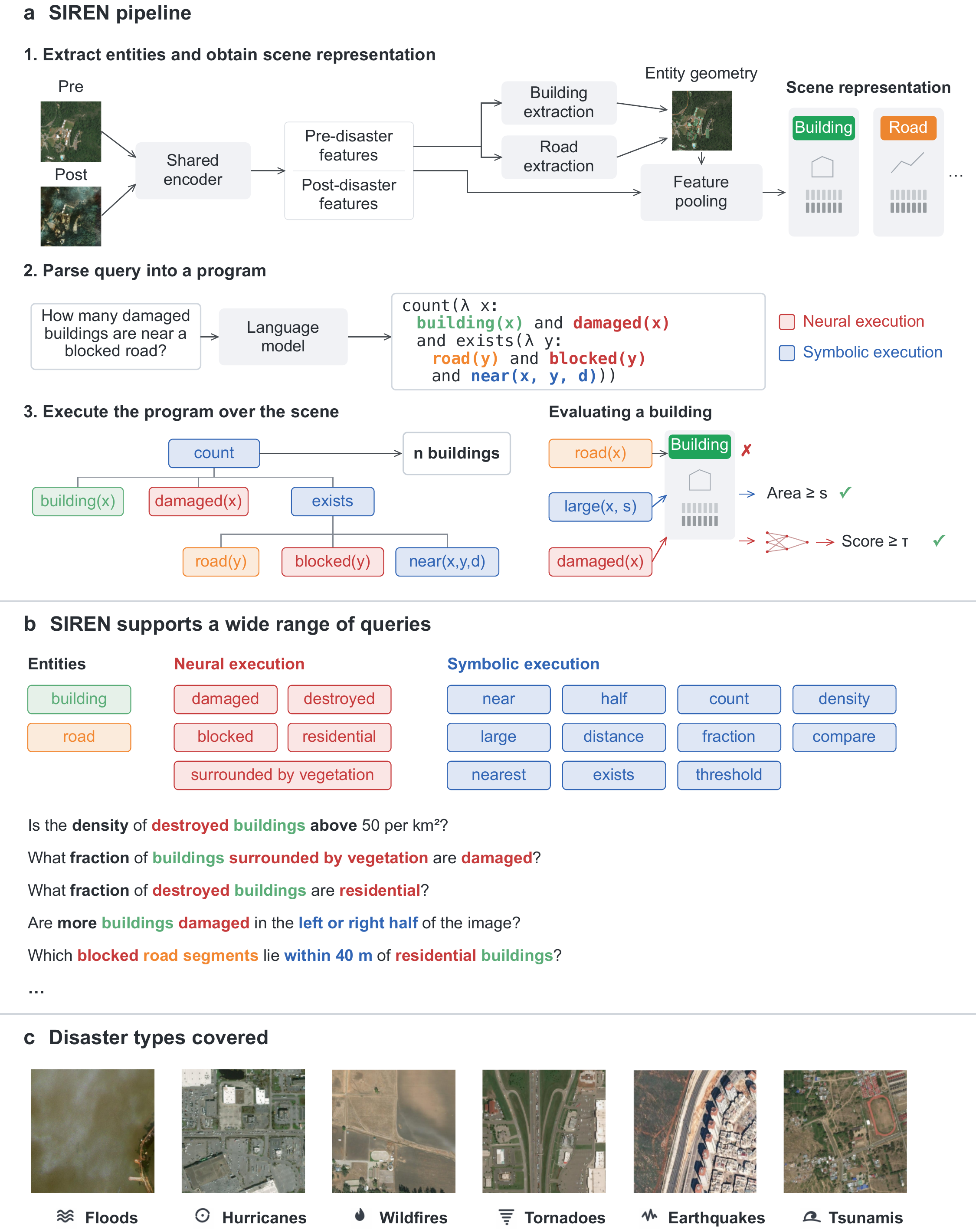}
\caption{SIREN combines neural and symbolic execution to answer disaster-response queries.
(a) Pipeline overview.
Buildings and roads are extracted from pre-disaster imagery and represented using shared pre- and post-disaster features.
Neural execution predicts visual properties such as damage; symbolic execution evaluates geometry and combines predictions to compute answers.
(b) A shared library of classifiers and functions supports a wide range of disaster-response queries.
(c) Examples of the six disaster types across the 15 DisasterM3 events.}
\label{fig:siren_overview}
\end{figure}

We compare SIREN with two categories of baselines: foundation models pretrained on satellite or aerial imagery, and frontier vision-language models.
The first category includes SatlasPretrain \cite{bastani2023satlaspretrain}, RemoteCLIP \cite{liu2024remoteclip}, SkyCLIP \cite{wang2023skyscript} and the satellite-pretrained version of DINOv3 \cite{simeoni2025dinov3}.
Each is fine-tuned separately to predict the count for one query, updating both its encoder and prediction head.
The second category comprises Claude Opus 5.5 and GPT-6.1 Sol, prompted to answer the same queries with explicit definitions of the relevant properties and conditions but without task-specific training.

\begin{figure}[p]
\centering
\includegraphics[width=\textwidth,height=0.86\textheight,keepaspectratio]{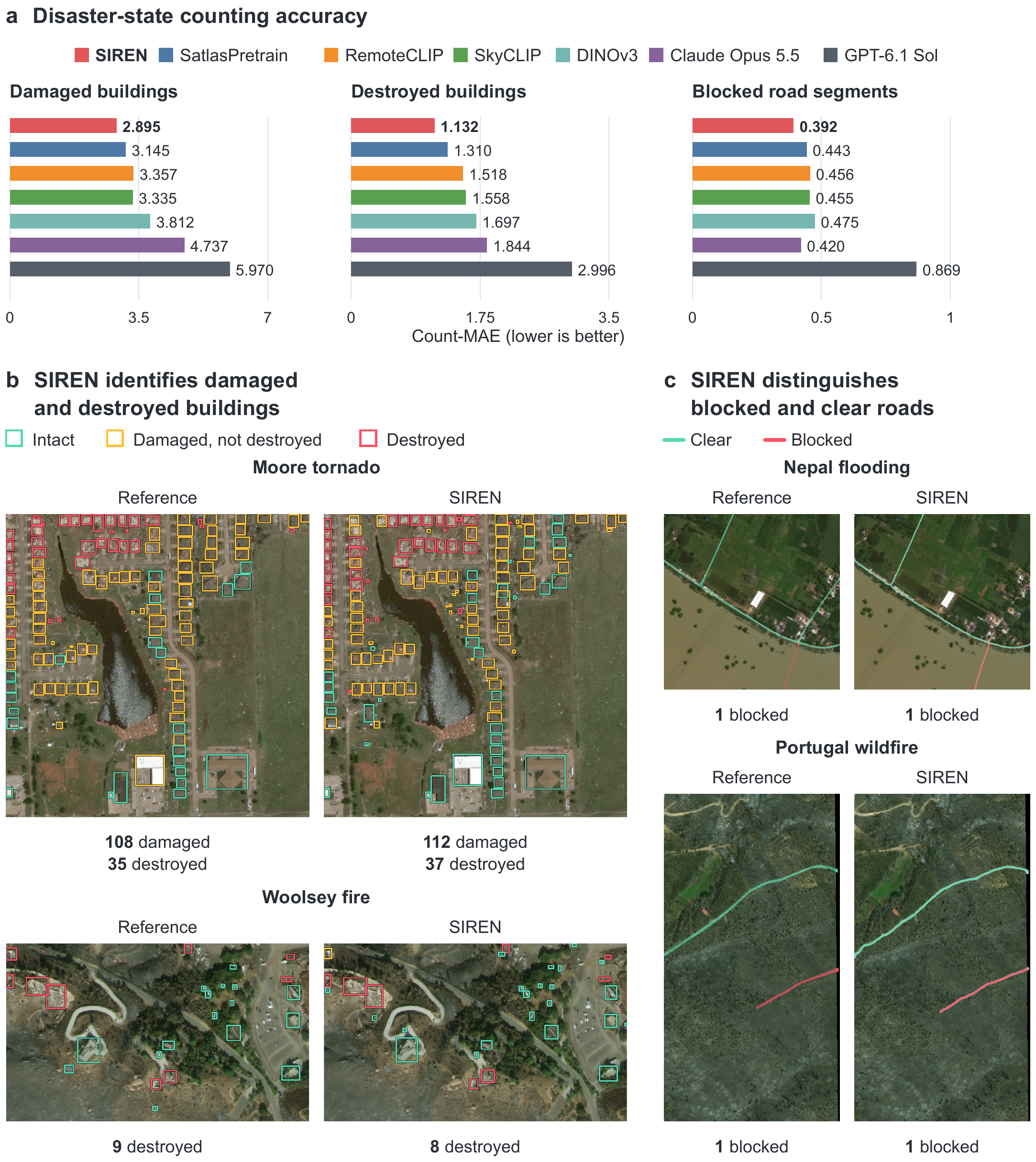}
\caption{SIREN recovers building and road damage states to answer disaster-response counting queries.
(a) Mean absolute counting error (Count-MAE) compared with the six baselines on the same eligible test scenes.
SIREN achieves the lowest error on all three queries.
Lower is better; counts of damaged buildings include destroyed buildings.
(b,c) Example predictions alongside reference labels show building states in tornado and wildfire scenes (b) and road conditions in flood and wildfire scenes (c).
Full numerical results are provided in Supplementary Table~1.}
\label{fig:core_results}
\end{figure}

SIREN achieves lower counting error than all six baselines on all three disaster-state queries (Fig.~\ref{fig:core_results}a; Supplementary Table~1).
Relative to the strongest baseline for each query, it reduces error by 8.0\% for damaged buildings, 13.6\% for destroyed buildings and 6.8\% for blocked road segments.
Example predictions show building and road damage states alongside reference labels across tornado, wildfire and flood scenes (Fig.~\ref{fig:core_results}b,c).

\subsection{SIREN supports compositional disaster-response analysis}

Beyond counting damaged buildings and blocked roads, disaster assessment often requires answering more specific queries about affected areas.
For example, counting destroyed residential buildings can help assess housing losses, while identifying damaged buildings without a nearby clear road can help prioritize areas where access may be difficult.

SIREN answers these queries using two types of execution.
First, learned classifiers predict properties such as building damage, road blockage, residential use and surrounding vegetation from the images (neural execution).
These predictions are associated with individual buildings and roads.
Second, a program uses the classifier predictions to answer the query (symbolic execution).
Different programs can combine the same classifier predictions with geometric calculations to answer different queries.
New combinations therefore require no additional training or query-specific labels.

\begin{figure}[p]
\centering
\includegraphics[width=\textwidth,height=0.86\textheight,keepaspectratio]{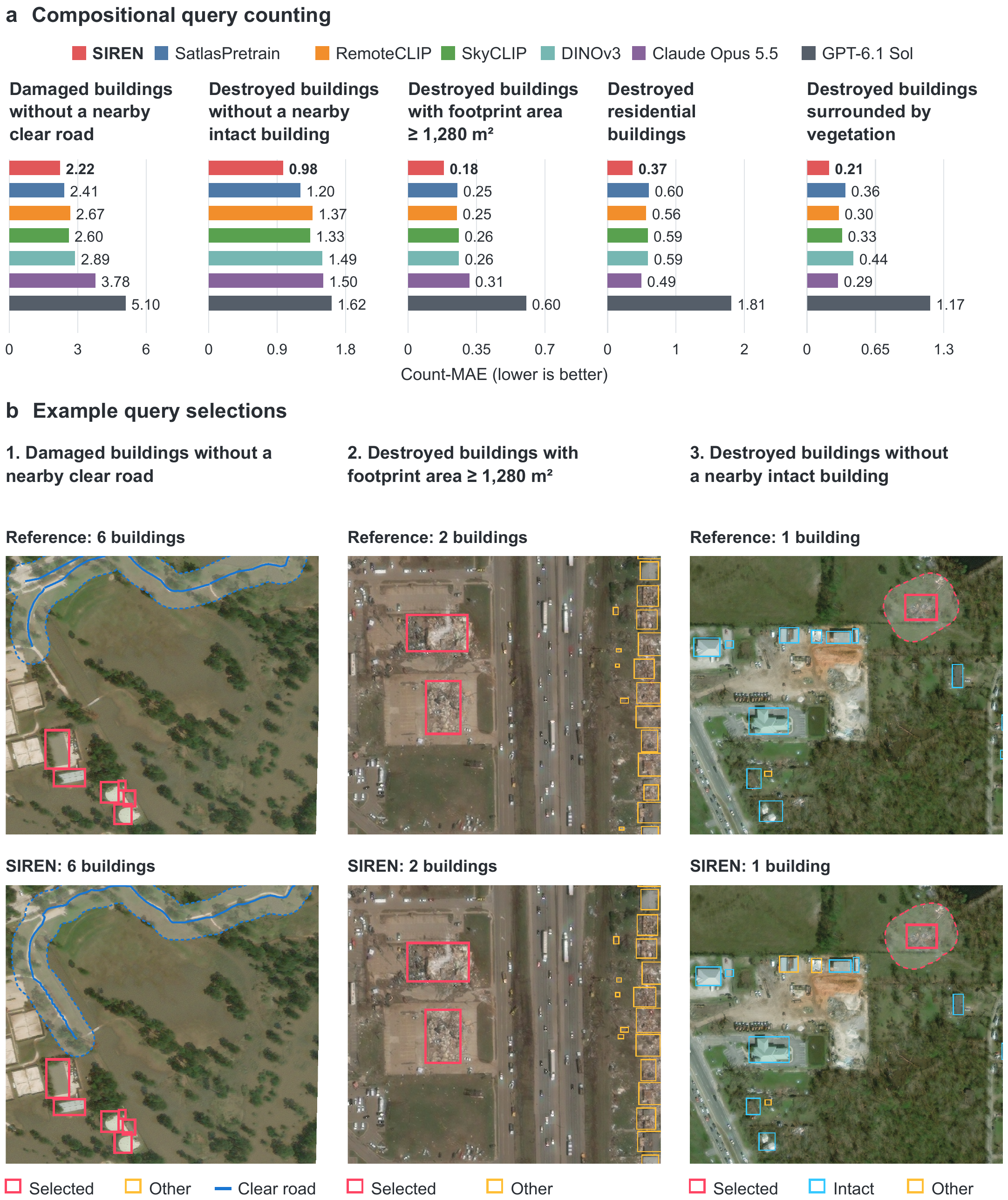}
\caption{SIREN answers compositional disaster-response queries.
(a) Count-MAE for five representative queries, comparing SIREN with the six baselines.
All methods use the same eligible scenes for each query; lower is better.
(b) SIREN's building selections alongside reference selections for three queries.
Full numerical results and query definitions are provided in Supplementary Table~2.}
\label{fig:compositional_queries}
\end{figure}

We demonstrate this capability on five representative disaster-response queries that combine damage with residential use, footprint area, surrounding vegetation or proximity to clear roads or intact buildings.
SIREN achieves the lowest counting error on all queries, reducing error by 7.7--28.9\% relative to the strongest baseline for each query (Fig.~\ref{fig:compositional_queries}a; Supplementary Table~2).
For example, for destroyed residential buildings, SIREN reduces error by 25.8\% compared with Claude, the strongest baseline for this query (0.366 versus 0.493).
Figure~\ref{fig:compositional_queries}b shows SIREN's building selections alongside reference selections for three representative queries.

\begin{figure}[p]
\centering
\includegraphics[width=\textwidth,height=0.86\textheight,keepaspectratio]{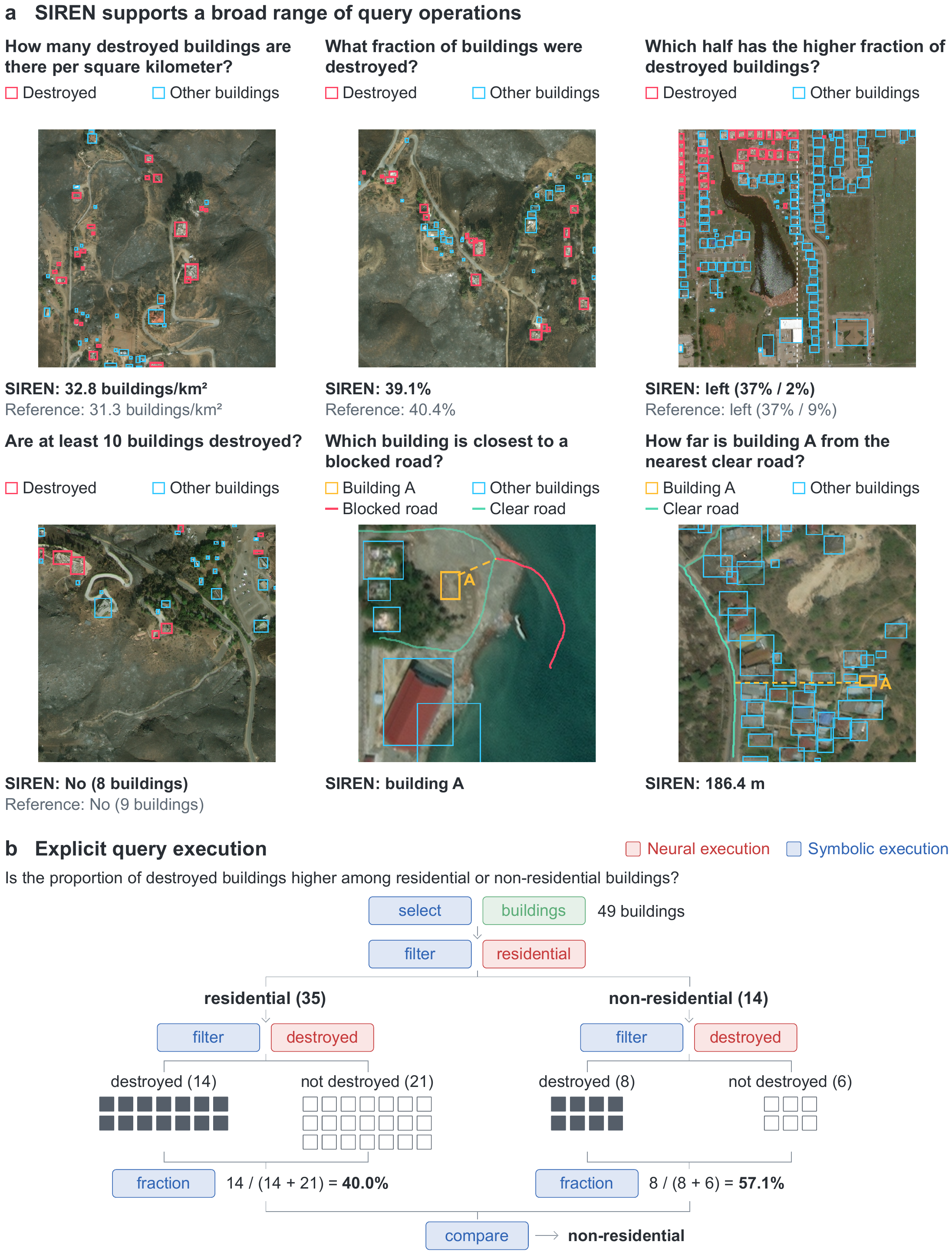}
\caption{SIREN supports a broad range of disaster-response queries.
(a) Examples of density, proportion, comparison, threshold, nearest-neighbor and distance queries.
(b) An example of explicit query execution, linking the answer to learned predictions and symbolic calculations that can be inspected step by step.}
\label{fig:query_breadth}
\end{figure}

SIREN's flexibility also extends beyond selecting and counting entities to performing different types of calculations (Fig.~\ref{fig:query_breadth}a).
For example, a program can separate residential and non-residential buildings, calculate the proportion destroyed in each group and compare the results (Fig.~\ref{fig:query_breadth}b).
Because the calculation is explicitly defined, the selected buildings and intermediate values can be inspected to trace how the result was obtained.
The same approach supports a wider range of analyses, including measuring damage density, comparing damage across regions, checking whether damage exceeds specified thresholds and calculating distances to nearby infrastructure.

\subsection{SIREN is computationally efficient}

Disaster assessments are time-critical: rapid-mapping services aim to deliver damage maps within hours of receiving imagery \cite{cems2021flyer}, and responders' information needs change as a response progresses from rescue to relief and recovery \cite{gralla2013workshop}.
We therefore compare SIREN with the baselines on both counting accuracy and the computational cost of answering the eight disaster-response queries.
We measure the total model computation and parameters required to answer all eight queries for an image pair.

SIREN analyzes each image pair once, extracting buildings and roads and predicting their properties.
It then answers queries by executing programs over this stored representation.
The baselines instead process the images separately for each query, using either a remote-sensing foundation model fine-tuned for that query or a request to a frontier vision-language model.

\begin{figure}[p]
\centering
\includegraphics[width=\textwidth,height=0.86\textheight,keepaspectratio]{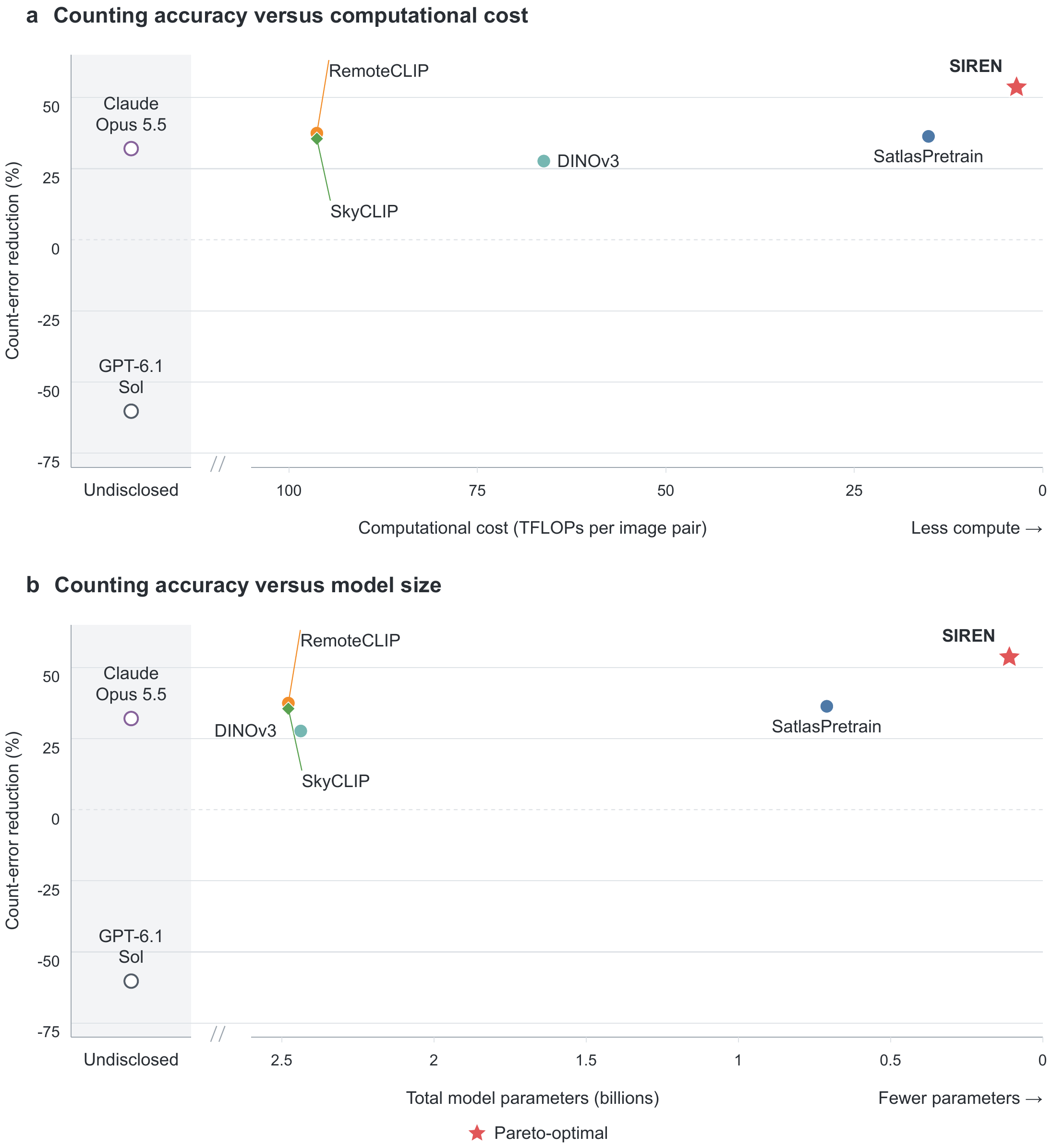}
\caption{SIREN combines accurate counting with efficient computation and parameter reuse.
Counting performance versus (a) computational cost and (b) total model parameters across all basic and compositional queries.
Performance is the mean per-query reduction in Count-MAE relative to predicting zero, evaluated on the same eligible test scenes for each query across all methods.
Stars mark Pareto-optimal systems among those with known resource measurements.
Vision-language models with undisclosed computation or parameter counts appear in unscaled columns.
Measurement details and numerical results are provided in Supplementary Table~3.}
\label{fig:efficiency}
\end{figure}

Across eight queries, SIREN achieves a higher count-error reduction than all six baselines (Fig.~\ref{fig:efficiency}; Supplementary Table~3).
SIREN uses 3.47 trillion floating-point operations (TFLOPs) per image pair, compared with 15.15 for SatlasPretrain, 66.19 for DINOv3 and 96.29 for RemoteCLIP and SkyCLIP.
It achieves lower error using 4.4--27.7 times less model computation than the remote-sensing baselines (Fig.~\ref{fig:efficiency}a).

SIREN also requires fewer total parameters: 0.11 billion, compared with 0.71--2.48 billion for each remote-sensing baseline, which needs a separate fine-tuned model for each query (Fig.~\ref{fig:efficiency}b).
It is the fastest: answering eight queries for an image pair takes 0.64 s, compared with 1.67--9.82 s for the remote-sensing baselines and 35.8 s and 98.0 s for Claude Opus 5.5 and GPT-6.1 Sol (Supplementary Table~3).
At DisasterM3's reported resolution, answering these queries for a 100\,km\textsuperscript{2} region would take about 1.6 minutes with SIREN, compared with 4--24 minutes for the remote-sensing baselines and about 1.5 and 4 hours for Claude Opus 5.5 and GPT-6.1 Sol.
By comparison, operational services in which analysts visually interpret satellite imagery deliver damage maps within hours or days \cite{copernicusems2026}.

\subsection{SIREN learns from limited annotations and transfers to new disasters}

\begin{figure}[p]
\centering
\includegraphics[width=\textwidth,height=0.86\textheight,keepaspectratio]{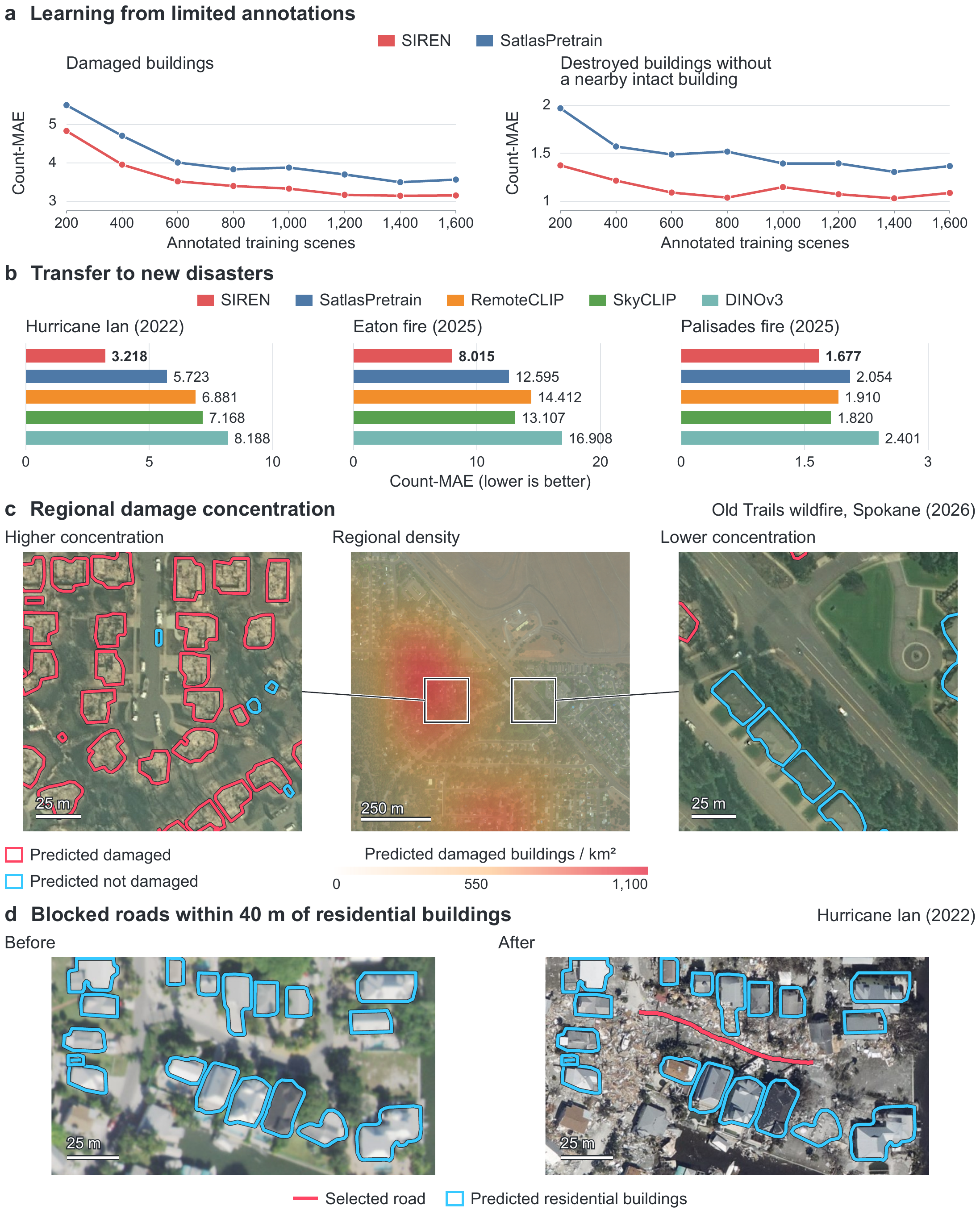}
\caption{SIREN learns from limited damage annotations and transfers to new disasters.
(a) Count-MAE for a basic and a compositional query as the number of annotated training scenes increases.
(b) Count-MAE for destroyed buildings following Hurricane Ian and the Eaton and Palisades fires.
(c) Regional density of buildings predicted as damaged or destroyed following the Old Trails wildfire in Spokane (2026).
(d) Predicted blocked roads within 40 m of residential buildings following Hurricane Ian (2022).
External analyses (b--d) use no event-specific training or calibration.
Numerical results are provided in Supplementary Tables~4 and~6.}
\label{fig:eaton_deployment}
\end{figure}

Beyond computational cost, disaster assessment is often constrained by limited training data, because building-level damage labels require expert annotation \cite{gupta2019xbd} and take time to collect after a new disaster.
We therefore examine how efficiently SIREN learns from limited annotations and whether it can assess new disasters without event-specific training.

We demonstrate annotation efficiency on two queries: counting damaged buildings and counting destroyed buildings without a nearby intact building.
The second adds a spatial condition, allowing us to examine whether SIREN's advantage is greater for compositional analysis.
We compare SIREN with SatlasPretrain, the strongest baseline on these queries in the main evaluation, using progressively larger subsets of the same damage-annotated training scenes, from 25 scenes to all training scenes.
For each subset, SIREN trains only its damage and destruction classifiers and keeps its building and road extraction unchanged, whereas SatlasPretrain is fine-tuned on the subset's counts for each query.

SIREN achieves lower error at every evaluated training-set size, with generally larger relative improvements in the compositional setting (Fig.~\ref{fig:eaton_deployment}a; Supplementary Table~4).
On the compositional query, SIREN achieves lower error with one quarter as many annotated scenes: 1.215 with 400 scenes, compared with 1.367 for SatlasPretrain with 1,600 scenes.
The same holds against all four remote-sensing baselines (Supplementary Table~5).
Overall, SIREN needs fewer annotated scenes than fine-tuned baselines to reach the same accuracy, especially for compositional queries.

We next apply SIREN and the remote-sensing baselines to \textit{unseen} disasters without event-specific training or calibration.
Transfer is challenging because imagery, building styles and the visual appearance of damage can vary across events.
For example, destroyed buildings may appear as burned ruins after wildfires or as collapsed structures and debris after hurricanes.
We count destroyed buildings after Hurricane Ian (2022), using aerial imagery, and the Eaton and Palisades fires (2025), using satellite imagery.
Reference counts come from independent structure-level damage records \cite{figueira2025ian,calfire2025eaton,calfire2025palisades}.
SIREN achieves the lowest counting error on all three events, reducing error by 43.8\%, 36.4\% and 7.9\%, respectively, relative to the strongest baseline (Fig.~\ref{fig:eaton_deployment}b; Supplementary Table~6).
These results show that SIREN can support rapid assessment of new disasters, including with imagery from different sources.

The same predictions also support other analyses in new disasters without additional training (Fig.~\ref{fig:eaton_deployment}c,d).
After the Old Trails wildfire in Spokane (2026), SIREN combines building predictions across scenes to map where damage is concentrated.
After Hurricane Ian, it finds roads predicted to be blocked within 40 m of residential buildings.

\FloatBarrier

\section{Discussion}

We present SIREN as a neurosymbolic framework for answering diverse disaster-response queries from satellite imagery.
Across different kinds of queries, SIREN achieves higher accuracy than state-of-the-art remote-sensing foundation models and frontier vision-language models, while requiring less computation.

Beyond accuracy and efficiency, SIREN provides flexibility as responders' priorities change from assessing initial damage to planning relief and recovery.
Rather than requiring a separate model for each query, SIREN learns reusable visual properties that can be combined in different ways.
Responders can therefore obtain new information from the same imagery without additional training or repeated visual processing.

As queries become more complex, understanding how an answer was obtained becomes especially important.
SIREN's explicit program execution makes the selected entities and calculations available for inspection.
An unexpected result can therefore be examined to determine whether it arose from a missed building, an incorrect classification or a condition specified in the query.
Crucially, these steps are the computation that produced the answer, not an explanation generated afterward.
They provide a basis for reviewing results and, in future work, correcting predictions before repeating an analysis.

The practical value of these analyses also depends on how readily they can be applied to a new disaster.
Responders need timely estimates, but annotated imagery may be scarce or unavailable early in a response.
We show that SIREN learns effectively from limited annotations and transfers to three new disasters without event-specific training or calibration.

A few limitations remain.
First, SIREN requires entity-level annotations that are more detailed than scene-level counts, although the resulting classifiers can support multiple queries.
Extending the system to new visual properties requires additional annotations for training classifiers.
Second, errors in entity extraction, classification and geometry can propagate into multiple answers.
Explicit execution makes the underlying predictions available for review, but does not eliminate their errors.
Finally, our quantitative evaluation focuses on counting queries, while broader symbolic operations are demonstrated through examples.
The evaluated programs are also written directly rather than generated from natural-language queries.

Future work could extend SIREN in three directions.
First, SIREN could be tested in real disaster response, where queries would be written in natural language and programs generated automatically.
Second, additional classifiers could cover more of the properties relevant to disaster response, such as flooded buildings, as annotations for them become available.
Third, SIREN could be adapted to conditions in which satellite views are limited, for example by using radar imagery when clouds or smoke obscure optical images.
Because SIREN's answers are estimates, they should support rather than replace expert assessment, particularly when they guide where resources are sent.

Ultimately, disaster responders need answers to many different queries, often with little time or local training data.
SIREN makes these analyses possible within a single, efficient and interpretable system.

\FloatBarrier

\section{Methods}

\subsection{Model architecture}

SIREN answers a disaster-response query in three steps (Fig.~\ref{fig:siren_overview}a).
First, it converts the pre- and post-disaster images into a scene representation: a set of buildings and road segments, each with its geometry and image features.
Second, it represents the query as a program, a sequence of simple operations on these entities.
Third, it executes the program on the scene representation, using learned classifiers to predict visual properties of entities (neural execution) and code to evaluate geometric conditions and calculations (symbolic execution).

\subsubsection{Scene representation}

Each image pair, or scene, shows the same geographic area before and after the disaster.
SIREN represents a scene in three steps: it encodes both images, locates buildings and roads, and then extracts features for each of them.
We refer to each building and road segment as an entity.

To encode both images, a shared SatlasPretrain Aerial Swin-B encoder with a feature pyramid network converts each image into feature maps at multiple spatial scales \cite{bastani2023satlaspretrain,liu2021swin,lin2017feature}.
Buildings and roads are then located from the pre-disaster feature maps, so their geometry does not depend on damage visible after the disaster.
A building detector, implemented as a Mask R-CNN instance segmentation head, predicts a mask and bounding box for each building \cite{he2017mask}.
We run it on the full image and on enlarged tiles in densely built areas to improve detection of small buildings, then remove duplicates and use the remaining masks to obtain building footprints, bounding boxes and areas.
A road segmentation head, implemented as a UPerNet semantic segmentation head, predicts a road mask \cite{xiao2018unified}.
We threshold and clean this mask, reduce it to a centerline network and divide the network into road segments, each with a bounding box.

Finally, for each entity, we use its bounding box, expanded to include the surrounding area, to pool the feature maps of both images into a fixed-size grid of 7~$\times$~7 cells with 128 channels per image (feature pooling in Fig.~\ref{fig:siren_overview}a).
Together with each entity's type and geometry (a footprint polygon and area for buildings, and a centerline and length for road segments), these features form the scene representation.
Supplementary Section~1 gives the extraction and pooling settings.

\subsubsection{Query programs}

Given a query, SIREN represents it as a program that is evaluated on the scene representation.
A program is a sequence of operations drawn from a shared library (Fig.~\ref{fig:siren_overview}b).
Each operation takes the results of earlier operations as inputs, such as a set of entities or a number, and produces a new result; the last operation gives the answer.
Because each operation can only use earlier results, SIREN can check that a program is valid before running it and can record the result of every operation.

For the quantitative evaluations, we use fixed programs written directly.
For queries written in natural language, a language model can generate a program from the same library.
The model receives the query text together with descriptions of the entity types, visual properties and operations, and returns a program in a fixed format, or no program if the query cannot be expressed with the library.
Supplementary Section~2 gives the model and generation settings.

\subsubsection{Neural execution}

Neural execution uses learned classifiers to predict visual properties of entities.
For each property, a classifier takes an entity's pooled features and outputs the probability that the entity has the property.
We train and evaluate five classifiers, which predict whether a building is damaged, destroyed, residential or surrounded by vegetation, and whether a road segment is blocked.

The damage and destruction classifiers compare the two images directly: they are small convolutional networks, with three convolutional layers, applied to the pre-disaster features, the post-disaster features and their difference.
The blocked-road classifier is a multilayer perceptron that also uses both images.
The residential and vegetation classifiers are multilayer perceptrons that use only pre-disaster features, because these properties describe the scene before the disaster.
The vegetation classifier pools features over a larger area around each building to include nearby trees and shrubs.

Each classifier is applied once to every relevant entity, and its probabilities are stored with the entities.
When a program uses a property, an entity has the property if its probability is at least a threshold selected on validation data.
New queries therefore reuse the stored predictions, and a new property requires only one additional classifier.
Supplementary Section~3 gives the layers and sizes of each classifier, and Supplementary Table~7 lists the inputs, label sources and thresholds.

\subsubsection{Symbolic execution}

Symbolic execution uses code, without learned parameters, to evaluate geometric conditions, logic and arithmetic over the stored geometries and predictions.
Each query program is built from three kinds of operations.
Selection operations return a subset of entities: those of a given type, those with or without a predicted property, the largest or smallest building, those in one half of the image, or those that are, or are not, within a given distance of entities in another set.
Measurement operations return the nearest entities between two sets and the distance between them.
Aggregation operations return a number from a set, such as its size, its fraction of a larger set or its density per square kilometer, and compare numbers with a threshold or with each other.
Operations can be combined in any order consistent with their input and output types, so the library supports many queries beyond those evaluated here.
New operations can be added by defining their input and output types, without retraining any model or recomputing stored predictions.

The distance between two entities is the shortest distance between their geometries, using footprints for buildings and centerlines for road segments.
In the evaluated queries, two entities are nearby if this distance is at most 50 pixels (40 m at DisasterM3's reported resolution of 0.8 m per pixel \cite{wang2025disasterm3}), and a building is large if its footprint covers at least 2,000 square pixels (1,280 m$^{2}$).
Other properties combine predictions: a building is intact if it is predicted to be neither damaged nor destroyed, and a road segment is clear if it is not predicted to be blocked.

The program for damaged buildings with no nearby clear road, for example, selects the buildings, keeps those predicted to be damaged, removes those within 50 pixels of a clear road segment and counts the rest.
Because the result of every operation is recorded, including which entities were selected, each answer can be traced to the entities and operations that produced it.
Supplementary Section~2 gives the program format, validation checks and calculation details.

\subsection{Datasets and annotations}

\subsubsection{DisasterM3}

DisasterM3 \cite{wang2025disasterm3} contains satellite imagery from 36 historical disaster events.
We use 7,711 RGB image pairs from 15 of these events that provide paired optical images with building and road annotations: five wildfires, four hurricanes, two floods, two tornadoes, one earthquake and one tsunami.
Supplementary Section~4 gives the selection criteria.
Each image contains 1024~$\times$~1024 pixels and covers about 820 m by 820 m at the dataset's reported resolution of 0.8 m per pixel \cite{wang2025disasterm3}.
We construct a custom split for evaluation, with 4,627 training, 1,157 validation (1,156 usable) and 1,927 test pairs, stratified by event and damage presence. 

DisasterM3 provides labels for damaged buildings, destroyed buildings and blocked roads.
The damaged category includes minor damage, major damage and destruction.
The destroyed category includes destruction only.
Blocked roads include flooded and debris-covered roads.
For each counting query, eligible scenes are those with sufficient reference annotations to determine the count.
For the road queries, we exclude image pairs with incomplete road annotations and evaluate all methods on the remaining pairs (Supplementary Section~4).

\subsubsection{Residential and vegetation labels}

DisasterM3 does not label residential use or surrounding vegetation, so we obtain these labels separately.
Residential training labels come from building-level records in the National Structure Inventory (NSI 2022) \cite{usace2022nsi}, explicit OpenStreetMap building tags \cite{openstreetmap} and Federal Emergency Management Agency (FEMA) USA Structures \cite{femausastructures}, in that priority order (Supplementary Section~5).
The residential test reference combines OpenStreetMap building and land-use tags with model-assisted human annotation.

Vegetation labels were generated by GPT-5.4 using a fixed rubric and validated by human review.
Trees or shrubs must dominate the building's immediate surroundings in the pre-disaster image.
Grass, lawns and crops alone do not qualify.

\subsubsection{External case studies}

To test transfer, we use disasters that are not among the 15 DisasterM3 events used for training and evaluation, with imagery and damage records from other sources.
For the 2025 Eaton and Palisades fires, we use paired Maxar Open Data satellite images \cite{maxar2025opendata} and CAL FIRE Damage Inspection (DINS) records \cite{calfire2025eaton,calfire2025palisades}.
For Hurricane Ian in 2022, we pair pre-disaster National Agriculture Imagery Program (NAIP) aerial images \cite{naip2021} with post-disaster National Oceanic and Atmospheric Administration (NOAA) aerial images \cite{noaa2022ian}.
Building damage labels come from the DesignSafe PRJ-5700 assessment \cite{figueira2025ian,amini2025iandata}.
For regional damage mapping after the Old Trails wildfire in Spokane (2026), we use Vantor Open Data imagery \cite{vantor2026opendata} covering about 15 km$^{2}$ in and around the fire perimeter. 
We prepare external scenes on a 1024~$\times$~1024-pixel grid covering 312.5 m by 312.5 m (approximately 0.305 m per pixel).

\subsection{Training}

We train SIREN in two stages: first to extract buildings and roads, then to predict their visual properties.
In the first stage, we jointly train the shared encoder, building detector and road segmentation head using DisasterM3 building footprint and road annotations.
We initialize the shared encoder and building detector from a Mask R-CNN model trained to detect buildings on the same training split, starting from SatlasPretrain weights, and train the road segmentation head from random initialization.
In the second stage, we freeze these components and extract entity features.
We train the classifiers on the stored features to predict visual properties, using class-weighted cross-entropy.

We train the classifiers for building damage and destruction using both annotated and detected buildings.
Including detected buildings exposes the classifiers to the imperfect detections encountered at inference.
Detected buildings receive labels by matching their bounding boxes to annotated buildings.
The other classifiers are trained on annotated buildings or road segments.
For each classifier, we retain the weights from the epoch with the highest validation macro-F1, the mean F1 score across the positive and negative classes.

We select all decision thresholds on validation data.
Thresholds for counting damaged buildings, destroyed buildings and blocked roads minimize Count-MAE; those for residential use and surrounding vegetation maximize macro-F1.
Compositional queries reuse these thresholds, except for the road-access query (damaged buildings without a nearby clear road), which uses the road threshold that maximizes macro-F1 on annotated road segments, to classify nearby roads rather than count them.
Supplementary Table~7 lists the selected thresholds.

Supplementary Section~6 gives the optimization settings and the rule for labeling detector outputs.

\subsection{Baselines}

We compare SIREN with two categories of baselines, both of which answer each query directly from the image pair rather than through an explicit program.

\subsubsection{Remote-sensing foundation models}

We build query-specific baselines from SatlasPretrain Aerial Swin-B, the ViT-L/14 versions of RemoteCLIP and SkyCLIP, and satellite-pretrained DINOv3 ViT-L/16 \cite{bastani2023satlaspretrain,liu2024remoteclip,wang2023skyscript,simeoni2025dinov3}.
Each model combines features from the pre- and post-disaster images and uses a prediction head to estimate a count.
For each query, we fine-tune a separate encoder and prediction head together using image-level counts, minimizing absolute error between predicted and reference counts.
Model weights are selected using the mean absolute counting error on validation scenes.

We use 1024~$\times$~1024-pixel inputs for SatlasPretrain and DINOv3.
For RemoteCLIP and SkyCLIP, we use 1022~$\times$~1022-pixel inputs to accommodate their 14-pixel patch size and interpolate positional embeddings to the input patch grid.
Supplementary Section~7 gives the architecture of the prediction head and training settings.

\subsubsection{Frontier vision-language models}

The second category of baselines comprises two frontier vision-language models, Claude Opus 5.5 and GPT-6.1 Sol.
Each receives paired pre- and post-disaster images and one query.
The prompt defines the relevant properties and geometric conditions.
Both models return a structured count without task-specific training.

For Claude, we compared requesting the count directly, listing qualifying entities before counting, and providing three labeled examples.
The direct-count prompt achieved the lowest aggregate validation counting error, so we use it for both models.
We evaluate Claude at high reasoning effort and GPT at low effort, which had the lowest combined validation Count-MAE for damaged and destroyed buildings among low, medium and high effort (Supplementary Table~8).

\subsection{Evaluation}

\subsubsection{Counting}

We report mean absolute counting error (Count-MAE), the mean absolute difference between predicted and reference counts per image pair.
We evaluate all models in each comparison on the same image pairs and reference answers.

To summarize counting error across eight queries, we divide each query's Count-MAE by that of always predicting zero and average these ratios with equal weight.
Figure~\ref{fig:efficiency} displays the count-error reduction, $100\times(1-\bar{e})$, where $\bar{e}$ is this mean normalized counting error.
A negative value indicates higher normalized error than predicting zero.

\subsubsection{Computational cost}

We measure the neural network computation required to answer all eight counting queries on 300 test image pairs, using the full-resolution configurations described above.
We report the mean floating-point operations (FLOPs) per image pair, including SIREN's additional tile passes for building detection, and count a multiply-accumulate as two operations.
Neural network operations are counted using PyTorch's FlopCounterMode.
This excludes CPU query execution, post-processing not captured by the profiler and optional program generation by a language model.

We report the total parameters used for inference by one shared SIREN system or eight query-specific models per remote-sensing baseline.
Figure~\ref{fig:efficiency} and Supplementary Table~3 combine these measurements with count-error reductions on the eligible test scenes.
API computation and parameter counts are undisclosed.
We also measure the time to answer all eight queries per image pair, using one NVIDIA RTX A5000 GPU with images already loaded for SIREN and the remote-sensing baselines, and end-to-end API request times, excluding one-time image uploads, for the vision-language models.

\subsubsection{Annotation efficiency}

For the two annotation-efficiency queries, SIREN and SatlasPretrain are trained on progressively larger, matched training subsets, from 25 scenes to all 4,627 scenes.
Supplementary Section~8 gives the subset sizes, sampling procedure and training schedules.

For SIREN, we keep the encoder and entity extractors trained on the full training split fixed and train only the damage classifiers on each annotated subset.
For SatlasPretrain, we fine-tune a separate encoder and prediction head for each query using counts derived from the same scenes.
We evaluate all models on the same 1,927 test scenes.
For each classifier trained on a subset, we select a decision threshold by minimizing Count-MAE for damaged or destroyed buildings on the full available validation set.
We keep these thresholds fixed for test evaluation, including the compositional query.

All methods use the same sequence of training subsets.
Figure~\ref{fig:eaton_deployment}a shows subsets of 200 to 1,600 scenes, and Supplementary Table~4 reports all sizes.
We also evaluate RemoteCLIP, SkyCLIP and DINOv3 on subsets of 100, 400 and 1,600 scenes from this sequence and on the full training set (Supplementary Table~5).

\subsubsection{External transfer}

For external transfer, SIREN counts detected buildings classified as destroyed.
The four remote-sensing baselines predict the total directly from the image pair.
Neither approach receives reference building locations.
We use model weights and decision thresholds selected using DisasterM3 data, without training or threshold calibration on external events.

For each scene, the reference count is the number of independent damage records within its boundaries labeled Destroyed in CAL FIRE DINS or the destroyed damage state (DS6) in the Hurricane Ian assessment.
We evaluate scenes containing at least one record, excluding Eaton scenes with an Inaccessible record because their destruction count cannot be determined.
The evaluation includes 101 scenes for Hurricane Ian, 131 for Eaton and 167 for Palisades.
We evaluate each image pair separately, including overlapping pairs from the fire events.

For the qualitative analyses in Fig.~\ref{fig:eaton_deployment}c,d, we use the same models and thresholds without event-specific training or calibration.
Supplementary Section~9 describes image preparation, regional density estimation and the road query.
Because no independent reference exists for the regional damage map or the road query, we present these analyses as examples and do not report accuracy metrics for them.

\subsection{Data availability}

DisasterM3 is available from its authors \cite{wang2025disasterm3}. The source imagery and reference records are available from the cited sources. Data splits, reference counts and saved predictions are included in the SIREN release at \url{https://github.com/arushig100/siren/releases/tag/v1.0.0}. Images in the figures are from DisasterM3 (CC BY-NC-SA 4.0), Vantor Open Data (CC BY-NC 4.0), NAIP and NOAA, and are cropped and overlaid with predictions or reference labels.

\subsection{Code availability}

The SIREN code is available at \url{https://github.com/arushig100/siren} under the MIT license. Trained model weights, installation instructions and a notebook with example images are included in release v1.0.0.

\bibliography{sn-bibliography}

\bmhead{Acknowledgements}
A.G. is supported by the National Science Foundation Graduate Research Fellowship Program under Grant No.~DGE-2146755.
A.G., J.H. and J.W. are in part supported by AFOSR YIP FA9550-23-1-0127, ONR YIP N00014-24-1-2117, Singapore MDDI AIVP-2025-008 and ONR MURI N00014-24-1-2748.

\end{document}


\clearpage
\setcounter{page}{1}
\renewcommand{\thepage}{S\arabic{page}}
\phantomsection
\pdfbookmark[0]{Supplementary information}{supplementary-information}
\begin{center}
{\Large\bfseries Supplementary information\par}
\vspace{6pt}
{\large Neurosymbolic disaster assessment from satellite imagery\par}
\end{center}
\vspace{6pt}
\setcounter{secnumdepth}{2}
\setcounter{section}{0}

\section*{Overview}
This supplement provides supporting details for the main paper, organized as follows.
\begin{itemize}
\item Section~\ref{supp:scene} describes how the scene representation is built: building detection, road extraction and feature pooling.
\item Section~\ref{supp:programs} describes how programs are generated and executed.
\item Section~\ref{supp:classifiers} describes the visual classifiers and how the damage classifiers were selected.
\item Section~\ref{supp:events} describes how disaster events were selected and which scenes are evaluated.
\item Section~\ref{supp:residential} describes how residential labels are matched to buildings.
\item Section~\ref{supp:training} gives the training settings for building and road extraction and for the classifiers.
\item Section~\ref{supp:baselines} describes the remote-sensing baselines.
\item Section~\ref{supp:learning} gives the settings for the annotation-efficiency experiments.
\item Section~\ref{supp:external} describes image preparation and spatial analyses for the external disasters.
\item Supplementary Tables~1--8 report detailed results and settings.
\end{itemize}

\setcounter{table}{0}
\renewcommand{\thetable}{\arabic{table}}
\renewcommand{\theHtable}{supp.\arabic{table}}
\renewcommand{\tablename}{Supplementary Table}

\section{Scene representation}\label{supp:scene}

We provide additional settings for the building detection, road extraction and feature pooling described in Methods.

\subsection{Building detection}

To improve detection of small buildings in densely built areas, we supplement full-image detection with detection on enlarged tiles.
We first run the detector on the full pre-disaster image.
If this produces at least 40 detections with confidence of at least 0.5, we divide the image into nine overlapping 512~$\times$~512-pixel tiles.
Each tile containing the centers of at least 25 of these detections is resized to 1024~$\times$~1024 pixels and processed by the detector again.

We merge the full-image and tile detections, retaining those with confidence of at least 0.4 and 0.3, respectively, and remove duplicates and buildings cut off at tile edges.
For scenes that do not meet the 40-detection criterion, we retain full-image detections with confidence of at least 0.5.

We selected these settings on validation data to reduce count error while limiting additional detector passes.
Applying tiling to every validation image improved counting in dense scenes but increased error in sparse scenes, motivating its selective use.

\subsection{Road extraction}

We convert the road segmentation head's pixel probabilities into individual road segments.
A threshold of 0.80, selected by maximizing validation F1 for road pixels, is applied, and small gaps are then filled and small fragments removed.
The resulting mask is reduced to a centerline network and divided at intersections.
We retain segments of at least 180 pixels to avoid counting short fragments as separate roads.

We apply the same procedure to the annotated road masks to define reference segments for training and evaluation.
A reference segment is labeled blocked if its centerline intersects a flooded or debris-covered road pixel.

\subsection{Feature pooling}

For each entity, we pool features over its bounding box and surrounding area.
We expand the box by 30 pixels on each side for building damage, destruction, residential use and road blockage.
For surrounding vegetation, we use a larger region to include nearby trees and shrubs: each side is expanded by the larger of 80 pixels and 0.75 times the box's longest side.
This matches the image area used to generate the vegetation labels.

We use RoIAlign to pool features from the feature-pyramid level matched to the region's size \cite{he2017mask,lin2017feature}, producing a 7~$\times$~7 grid with 128 channels per image.
All entity features come from the full-image feature maps, including those for buildings detected in enlarged tiles.

\section{Program generation and execution}\label{supp:programs}

We describe the program format, validation checks and implementation of the query examples introduced in Methods.

Programs are represented in JSON and contain at most 64 operations.
Before execution, SIREN checks that every operation is supported, that each input refers to an earlier result of the required type and that each visual property is defined for the entity type.
Invalid programs are rejected.
Execution includes only the operations needed to obtain the answer.

For the optional natural-language translation described in Methods, our implementation uses Claude Sonnet 4.6 with temperature 0 and an output limit of 2,048 tokens.

For the examples in Fig.~4, density is the number of selected entities divided by the region's area, and a proportion is the number selected divided by the number in the relevant group.
Regional comparisons apply the same calculation to each region.
Distances are the shortest separations between entity geometries: building comparisons use footprint polygons, and building-to-road comparisons use the road centerline.
A distance with no eligible entities or a proportion with an empty denominator is returned as undefined.

Distance and area conditions for DisasterM3 queries are defined in pixels.
The figures report them in meters using the dataset's reported resolution of 0.8 m per pixel \cite{wang2025disasterm3}; external analyses use their own georeferenced grids.
Queries always run on full scenes; crops in the figures affect only the display.

\section{Visual classifiers}\label{supp:classifiers}

We provide additional details on the visual classifier architectures and the validation comparison used to select the damage and destruction classifiers.

\subsection{Architectures}

We use convolutional networks for building damage and destruction, and multilayer perceptrons (MLPs) for road blockage, residential use and surrounding vegetation.

Each convolutional classifier concatenates the pre-disaster feature grid, the post-disaster feature grid and their signed difference (post minus pre), giving 384 input channels.
Three convolutional layers with 256 channels (1~$\times$~1, 3~$\times$~3 and 3~$\times$~3), each followed by batch normalization and ReLU, process this input.
The classifier concatenates the spatial mean and maximum of the resulting features, applies dropout of 0.3 and predicts the class through two fully connected layers with 256 hidden units.

Each MLP first reduces a flattened 7~$\times$~7 grid (6,272 values) to 64 values with a linear layer, ReLU and dropout.
The road-blockage MLP receives the reduced pre- and post-disaster features and their absolute difference; the residential and vegetation MLPs receive the reduced pre-disaster features only.

\subsection{Selection of damage classifiers}

We compared three configurations for building damage and destruction: an MLP trained on annotated and detected buildings, a convolutional network trained on annotated buildings only, and a convolutional network trained on annotated and detected buildings.
We selected the last configuration because it achieved the lowest combined validation Count-MAE for counting damaged and destroyed buildings.

\section{Events and evaluation scenes}\label{supp:events}

We describe how events and eligible test scenes were selected for the DisasterM3 evaluations.

We require paired optical images with building and road annotations.
From the dataset's 36 events, we excluded those without road annotations, those with post-disaster radar (SAR) images or no matching pre-disaster images, and those with fewer than 100 image pairs.
We also excluded four smaller events whose disaster types were already represented (the Mexico earthquake, Joplin tornado, Sunda tsunami and Hurricane Ian).
The Lower Puna volcanic eruption was excluded because its lava-flow damage differs from that in the other events and its images lack the location metadata needed to derive residential labels.

Within these events, each query is evaluated on scenes with sufficient reference annotations to determine its answer.
All seven methods use the same eligible test scenes: 1,927 image pairs for counting damaged buildings, destroyed buildings, destroyed buildings without a nearby intact building and large destroyed buildings; 1,748 for blocked roads and damaged buildings without a nearby clear road; 1,372 for destroyed residential buildings; and 129 for destroyed buildings surrounded by vegetation.
Missing annotations are not treated as zero counts.

For the road queries, we also exclude Turkey earthquake test pairs with missing road annotations.
This exclusion applies only to evaluation; training data, trained models and decision thresholds remain unchanged.

\section{Residential label matching}\label{supp:residential}

We assign residential training labels by matching building footprints to the external records described in Methods.
Each building receives a label from the first source with a definite classification, in the order National Structure Inventory (NSI), OpenStreetMap and FEMA USA Structures.

We match NSI occupancy points to the footprint that contains them.
When several points fall within a footprint, we use the majority class and leave ties unresolved.
NSI occupancy types beginning with RES are residential; commercial, industrial, agricultural, religious, government and education types are non-residential.
For records with footprint geometries, we match by containment and then, where applicable, by the nearest footprint within 10 m.
We use current OpenStreetMap tags.
Buildings without a resolved label are excluded from classifier training.

\section{Training settings}\label{supp:training}

We provide optimization settings for the two training stages described in Methods and explain how detector outputs are labeled for classifier training.

\subsection{Building and road extraction}

We jointly train the shared encoder, building detector and road segmentation head with the Mask R-CNN detection losses and a road segmentation loss weighted by 0.2.
The road loss sums a binary cross-entropy term, in which road pixels are weighted by the ratio of non-road to road pixels in the batch (at most 50), a Dice term with weight 4 and a centerline Dice term with weight 2 \cite{shit2021cldice}.
Training uses images with at least one annotated building, and each loss uses only images with the corresponding annotations.

We train with AdamW for 10,000 optimizer updates, accumulating gradients over four batches of two images (effective batch size eight; 40,000 batches in total).
The learning rate is 0.000005 for the pretrained encoder and 0.0002 for the detection and segmentation heads.
Weight decay is 0.0001, the learning rate follows a cosine decay after 1,000 warm-up batches, and the final weights are retained.

\subsection{Classifiers}

For the damage and destruction classifiers, training includes both annotated buildings and detector outputs.
We collect detector outputs using less restrictive settings than at inference: all nine tiles are processed when the full image contains at least 30 detections with confidence of at least 0.5, and all candidates with confidence of at least 0.3 are retained, including duplicates.

Each detection receives the labels of the annotated building with the highest bounding-box intersection over union (IoU), provided that IoU is at least 0.5.
Detections with a maximum IoU below 0.2 are negative examples; the remaining detections are excluded.
During training, we apply the same random flips and transpositions to the pre- and post-disaster feature grids in each minibatch.

We train the classifiers for building damage and destruction separately for 20 epochs with a batch size of 512, and the other classifiers for 30 epochs with a batch size of 256.
All classifiers use AdamW with a learning rate of 0.001, weight decay of 0.0001 and cosine learning-rate decay, and cross-entropy class weights proportional to the inverse square root of class frequency.

\section{Remote-sensing baselines}\label{supp:baselines}

We provide details of the prediction head and fine-tuning settings for the four remote-sensing foundation models.

All four models use the same prediction-head architecture.
The head receives the concatenated pre- and post-disaster feature vectors and their absolute difference.
A two-layer network with 256 hidden units, ReLU activation and dropout of 0.1 maps this input to a count, which is rounded to a non-negative integer before scoring.

We fine-tune each encoder and prediction head for 25 epochs with AdamW, weight decay of 0.0001 and cosine learning-rate decay.
Both images in a pair receive the same spatial augmentation, and training balances image pairs with positive and zero reference counts.
For each model, these settings are shared across all eight queries.
The table below lists the settings that differ between models; batch sizes refer to image pairs.

{\small
\setlength{\tabcolsep}{3pt}
\renewcommand{\arraystretch}{1.15}
\noindent\begin{tabular}{>{\raggedright\arraybackslash}p{\dimexpr0.2\linewidth-2\tabcolsep\relax}>{\raggedright\arraybackslash}p{\dimexpr0.17\linewidth-2\tabcolsep\relax}>{\raggedright\arraybackslash}p{\dimexpr0.14\linewidth-2\tabcolsep\relax}>{\raggedright\arraybackslash}p{\dimexpr0.11\linewidth-2\tabcolsep\relax}>{\raggedright\arraybackslash}p{\dimexpr0.2\linewidth-2\tabcolsep\relax}>{\raggedright\arraybackslash}p{\dimexpr0.18\linewidth-2\tabcolsep\relax}}
\toprule
\textbf{Model} & \textbf{Input size} & \textbf{Learning rate} & \textbf{Batch size} & \textbf{Accumulation steps} & \textbf{Effective batch size} \\
\midrule
SatlasPretrain & 1024~$\times$~1024 & 0.0001 & 2 & 16 & 32 \\
DINOv3 & 1024~$\times$~1024 & 0.00001 & 1 & 32 & 32 \\
RemoteCLIP & 1022~$\times$~1022 & 0.00001 & 2 & 16 & 32 \\
SkyCLIP & 1022~$\times$~1022 & 0.00001 & 2 & 16 & 32 \\
\bottomrule
\end{tabular}
}

\section{Annotation-efficiency experiments}\label{supp:learning}

We describe subset sampling and training settings for the annotation-efficiency experiments.

Subset sizes are 25 and 100 scenes, every 200 scenes from 200 to 1,600, and all 4,627 training scenes.
We randomly order the training scenes once and use the first $N$ scenes for a subset of size $N$.
Smaller subsets are therefore contained in larger ones.
Both queries and all methods use the same scene lists.

SIREN's damage and destruction classifiers train for 20 epochs at each subset size.
For each classifier, we search decision thresholds from 0.30 to 0.90 in steps of 0.05, using the validation criterion described in Methods.
SatlasPretrain uses the same number of training steps as in full-data training, so reducing the subset size does not reduce its number of optimizer updates.

\section{External image preparation and spatial analyses}\label{supp:external}

We describe image preparation for the external disasters, followed by the regional density calculation and road query.

\subsection{Acquisition and alignment}

We prepare external image pairs on a common analysis grid, with 1024 pixels spanning 312.5 m (approximately 0.305 m per pixel).
This spacing does not imply that every source image has the same native resolution.
All methods receive these images without resampling to the DisasterM3 resolution.
Image sources and acquisition dates are listed below.

{\small
\setlength{\tabcolsep}{3pt}
\renewcommand{\arraystretch}{1.15}
\noindent\begin{tabular}{>{\raggedright\arraybackslash}p{\dimexpr0.22\linewidth-2\tabcolsep\relax}>{\raggedright\arraybackslash}p{\dimexpr0.43\linewidth-2\tabcolsep\relax}>{\raggedright\arraybackslash}p{\dimexpr0.35\linewidth-2\tabcolsep\relax}}
\toprule
\textbf{Event} & \textbf{Pre-disaster imagery} & \textbf{Post-disaster imagery} \\
\midrule
Eaton fire & Maxar, 25 December 2024 or 1 January 2025 & Maxar, 15 January 2025 \\
Palisades fire & Maxar, 21 December 2024 & Maxar, 15 January 2025 \\
Hurricane Ian & NAIP, 2021 & NOAA, 30 September 2022 \\
Old Trails wildfire, Spokane & Vantor Open Data (WorldView-3), 23 June 2026 & Vantor Open Data (Legion-1), 4 August 2026 \\
\bottomrule
\end{tabular}
}

We extract Eaton and Palisades scenes directly from Maxar's analysis-ready image tiles, which use this grid on both dates.
For Hurricane Ian, we resample both images to UTM zone 17N (EPSG:26917) with bilinear interpolation.
A translational correction for each post-disaster image is estimated using phase correlation and applied when the shift is at most 100 pixels along each axis.

We resample the Spokane images to UTM zone 11N (EPSG:32611) with bilinear interpolation.
The source acquisitions are B040001100002710 before the fire and B110001100F60310 after it.
Phase-correlation corrections are capped at 100 pixels per axis and kept only if they do not increase the measured residual shift.

We cut Spokane scenes from a fixed, non-overlapping grid, keeping cells with at least 25\% of their area inside the published NIFC Old Trails fire perimeter \cite{nifc2026perimeters} (12.9 km$^{2}$) and at least 95\% valid pixels on both dates.
This gives 155 scenes covering 15.1 km$^{2}$.

\subsection{Regional damage density}

For the Spokane regional map, we aggregate building predictions across the 155 scenes.
Each building predicted as damaged or destroyed contributes one footprint centroid to a grid with cells 32 image pixels wide (approximately 9.8 m).
We smooth the centroid counts with a Gaussian kernel with a standard deviation of 100 m.
To account for incomplete image coverage, we smooth the coverage mask with the same kernel and divide the smoothed counts by the smoothed coverage and grid-cell area.
The result is the predicted number of damaged or destroyed buildings per square kilometer.
We omit locations outside image coverage and do not deduplicate building fragments at scene boundaries.

\subsection{Hurricane Ian road query}

For Hurricane Ian, we select predicted blocked roads within 40 m of at least one predicted residential building, using the shortest distance between their geometries.
Road classification uses the road-access threshold listed in Supplementary Table~\ref{tab:neural_concepts}.

\clearpage
\section*{Supplementary tables}
\addcontentsline{toc}{section}{Supplementary tables}

{\small
\setlength{\tabcolsep}{4pt}
\renewcommand{\arraystretch}{1.12}
\begin{longtable}{>{\raggedright\arraybackslash}p{\dimexpr0.28\linewidth-8pt\relax}>{\raggedright\arraybackslash}p{\dimexpr0.24\linewidth-8pt\relax}>{\raggedright\arraybackslash}p{\dimexpr0.24\linewidth-8pt\relax}>{\raggedright\arraybackslash}p{\dimexpr0.24\linewidth-8pt\relax}}
\caption{Basic disaster-state counting}\label{tab:core_counting}\\*
\multicolumn{4}{p{\dimexpr\linewidth-8pt\relax}}{Count-MAE for SIREN, four remote-sensing foundation models and two frontier vision-language models on the same eligible test scenes. Lower is better. Error reductions are relative to the strongest baseline for each query.}\\*[5pt]
\toprule
Method & Damaged buildings & Destroyed buildings & Blocked road segments \\*
\midrule
\endfirsthead
\multicolumn{4}{l}{Supplementary Table~\thetable{} (continued)}\\
\toprule
Method & Damaged buildings & Destroyed buildings & Blocked road segments \\
\midrule
\endhead
\endfoot
\endlastfoot
SIREN & \textbf{2.895} & \textbf{1.132} & \textbf{0.392} \\*
SatlasPretrain & 3.145 & 1.310 & 0.443 \\*
RemoteCLIP & 3.357 & 1.518 & 0.456 \\*
SkyCLIP & 3.335 & 1.558 & 0.455 \\*
DINOv3 & 3.812 & 1.697 & 0.475 \\*
Claude Opus 5.5 & 4.737 & 1.844 & 0.420 \\*
GPT-6.1 Sol & 5.970 & 2.996 & 0.869 \\*
Always zero & 9.305 & 3.265 & 0.729 \\*
SIREN error reduction (\%) & 8.0 & 13.6 & 6.8 \\*
\bottomrule
\addlinespace[6pt]
\multicolumn{4}{p{\dimexpr\linewidth-8pt\relax}}{Damaged buildings include destroyed buildings.}\\*[5pt]
\end{longtable}
}

\bigskip
\needspace{0.55\textheight}
{\small
\setlength{\tabcolsep}{4pt}
\renewcommand{\arraystretch}{1.12}
\begin{longtable}{>{\raggedright\arraybackslash}p{\dimexpr0.23\linewidth-8pt\relax}>{\raggedright\arraybackslash}p{\dimexpr0.154\linewidth-8pt\relax}>{\raggedright\arraybackslash}p{\dimexpr0.154\linewidth-8pt\relax}>{\raggedright\arraybackslash}p{\dimexpr0.154\linewidth-8pt\relax}>{\raggedright\arraybackslash}p{\dimexpr0.154\linewidth-8pt\relax}>{\raggedright\arraybackslash}p{\dimexpr0.154\linewidth-8pt\relax}}
\caption{Compositional counting}\label{tab:compositional_counting}\\*
\multicolumn{6}{p{\dimexpr\linewidth-8pt\relax}}{Count-MAE for SIREN and the six baselines on the same eligible test scenes. Lower is better. Error reductions are relative to the strongest baseline for each query.}\\*[5pt]
\toprule
Method & Damaged, no nearby clear road & Destroyed, no nearby intact building & Large destroyed buildings & Destroyed residential buildings & Destroyed, surrounded by vegetation \\*
\midrule
\endfirsthead
\multicolumn{6}{l}{Supplementary Table~\thetable{} (continued)}\\
\toprule
Method & Damaged, no nearby clear road & Destroyed, no nearby intact building & Large destroyed buildings & Destroyed residential buildings & Destroyed, surrounded by vegetation \\
\midrule
\endhead
\endfoot
\endlastfoot
SIREN & \textbf{2.222} & \textbf{0.979} & \textbf{0.182} & \textbf{0.366} & \textbf{0.209} \\*
SatlasPretrain & 2.407 & 1.204 & 0.252 & 0.605 & 0.364 \\*
RemoteCLIP & 2.672 & 1.365 & 0.249 & 0.564 & 0.302 \\*
SkyCLIP & 2.605 & 1.333 & 0.258 & 0.587 & 0.333 \\*
DINOv3 & 2.887 & 1.494 & 0.260 & 0.592 & 0.442 \\*
Claude Opus 5.5 & 3.776 & 1.505 & 0.313 & 0.493 & 0.295 \\*
GPT-6.1 Sol & 5.098 & 1.616 & 0.604 & 1.810 & 1.171 \\*
Always zero & 5.900 & 2.730 & 0.444 & 0.591 & 0.279 \\*
SIREN error reduction (\%) & 7.7 & 18.7 & 26.9 & 25.8 & 28.9 \\*
\bottomrule
\addlinespace[6pt]
\multicolumn{6}{p{\dimexpr\linewidth-8pt\relax}}{Nearby means a minimum geometric separation of at most 50 pixels. For these queries, large buildings have a footprint area of at least 2,000 square pixels. A building is considered intact when both its damaged and destroyed probabilities are below their respective thresholds. Surrounding vegetation is assessed from pre-disaster imagery.}\\*[5pt]
\end{longtable}
}

{\small
\setlength{\tabcolsep}{4pt}
\renewcommand{\arraystretch}{1.12}
\begin{longtable}{>{\raggedright\arraybackslash}p{\dimexpr0.24\linewidth-8pt\relax}*{4}{>{\raggedright\arraybackslash}p{\dimexpr0.19\linewidth-8pt\relax}}}
\caption{Counting performance and computational resources}\label{tab:efficiency_current}\\*
\multicolumn{5}{p{\dimexpr\linewidth-8pt\relax}}{Performance and resources for the three basic and five compositional queries. Count-error reduction is the mean per-query reduction in Count-MAE relative to always predicting zero; higher is better. Computational cost, time and parameter totals cover all eight queries per image pair; lower is better.}\\*[5pt]
\toprule
Method & Count-error reduction vs zero (\%) & Computational cost (TFLOPs/image pair) & Time (s/image pair) & Total parameters (billions) \\*
\midrule
\endfirsthead
\multicolumn{5}{l}{Supplementary Table~\thetable{} (continued)}\\
\toprule
Method & Count-error reduction vs zero (\%) & Computational cost (TFLOPs/image pair) & Time (s/image pair) & Total parameters (billions) \\
\midrule
\endhead
\endfoot
\endlastfoot
SIREN & \textbf{53.64} & \textbf{3.470} & \textbf{0.64} & \textbf{0.110} \\*
SatlasPretrain & 36.35 & 15.153 & 1.67 & 0.709 \\*
RemoteCLIP & 37.47 & 96.293 & 9.77 & 2.478 \\*
SkyCLIP & 35.51 & 96.293 & 9.82 & 2.478 \\*
DINOv3 & 27.67 & 66.187 & 6.87 & 2.437 \\*
Claude Opus 5.5 & 32.05 & Undisclosed & 35.8 & Undisclosed \\*
GPT-6.1 Sol & $-$60.27 & Undisclosed & 98.0 & Undisclosed \\*
\bottomrule
\addlinespace[6pt]
\multicolumn{5}{p{\dimexpr\linewidth-8pt\relax}}{For each query, Count-MAE is divided by the error of always predicting zero. Count-error reduction is 100 times one minus the mean of these ratios, with all eight queries weighted equally. It is not classification accuracy; a negative value means higher normalized error than predicting zero.}\\*[5pt]
\multicolumn{5}{p{\dimexpr\linewidth-8pt\relax}}{FLOPs measure visual model computation, counting a multiply-accumulate as two operations. They exclude program generation by a language model, CPU query execution and post-processing not captured by the profiler. Parameter totals cover one shared SIREN system or eight separately fine-tuned baseline models.}\\*[5pt]
\multicolumn{5}{p{\dimexpr\linewidth-8pt\relax}}{Times for SIREN and the remote-sensing baselines are measured on one NVIDIA RTX A5000 GPU over 300 test image pairs, with images already loaded. Times for the vision-language models are measured at the client on up to 400 test pairs per query, summed over the eight requests, and include network and server latency.}\\*[5pt]
\end{longtable}
}

\bigskip
{\small
\setlength{\tabcolsep}{4pt}
\renewcommand{\arraystretch}{1.12}
\begin{longtable}{>{\raggedright\arraybackslash}p{\dimexpr0.16\linewidth-8pt\relax}*{4}{>{\centering\arraybackslash}p{\dimexpr0.21\linewidth-8pt\relax}}}
\caption{Learning with limited damage annotations}\label{tab:learning_dense}\\*
\multicolumn{5}{p{\dimexpr\linewidth-8pt\relax}}{Test Count-MAE for one sequence of training subsets shared by both methods, evaluated on the same 1,927 scenes; lower is better. The compositional query counts destroyed buildings without a nearby intact building (within 50 pixels).}\\*[5pt]
\toprule
& \multicolumn{2}{c}{Damaged buildings} & \multicolumn{2}{c}{Destroyed, no nearby intact building} \\*
\cmidrule(lr){2-3}\cmidrule(lr){4-5}
Training scenes & SIREN & SatlasPretrain & SIREN & SatlasPretrain \\*
\midrule
\endfirsthead
\multicolumn{5}{l}{Supplementary Table~\thetable{} (continued)}\\
\toprule
& \multicolumn{2}{c}{Damaged buildings} & \multicolumn{2}{c}{Destroyed, no nearby intact building} \\
\cmidrule(lr){2-3}\cmidrule(lr){4-5}
Training scenes & SIREN & SatlasPretrain & SIREN & SatlasPretrain \\
\midrule
\endhead
\endfoot
\endlastfoot
25 & \textbf{7.224} & 9.126 & \textbf{2.220} & 2.585 \\*
100 & \textbf{6.116} & 8.095 & \textbf{1.476} & 2.650 \\*
200 & \textbf{4.834} & 5.508 & \textbf{1.374} & 1.970 \\*
400 & \textbf{3.955} & 4.708 & \textbf{1.215} & 1.571 \\*
600 & \textbf{3.518} & 4.010 & \textbf{1.090} & 1.488 \\*
800 & \textbf{3.398} & 3.833 & \textbf{1.037} & 1.518 \\*
1,000 & \textbf{3.328} & 3.878 & \textbf{1.148} & 1.394 \\*
1,200 & \textbf{3.166} & 3.698 & \textbf{1.073} & 1.394 \\*
1,400 & \textbf{3.142} & 3.496 & \textbf{1.030} & 1.306 \\*
1,600 & \textbf{3.150} & 3.564 & \textbf{1.087} & 1.367 \\*
All 4,627 & \textbf{2.895} & 3.145 & \textbf{0.979} & 1.204 \\*
\bottomrule
\addlinespace[6pt]
\multicolumn{5}{p{\dimexpr\linewidth-8pt\relax}}{SIREN trains damage and destruction classifiers while keeping its encoder and entity extraction heads fixed after training on the full split without damage labels. SatlasPretrain is fine-tuned separately for each query. SIREN thresholds are selected on 1,156 validation scenes. Training and threshold selection are described in Methods.}\\*[5pt]
\end{longtable}
}

\bigskip
{\small
\setlength{\tabcolsep}{4pt}
\renewcommand{\arraystretch}{1.12}
\begin{longtable}{>{\raggedright\arraybackslash}p{\dimexpr0.26\linewidth-8pt\relax}>{\raggedright\arraybackslash}p{\dimexpr0.185\linewidth-8pt\relax}>{\raggedright\arraybackslash}p{\dimexpr0.185\linewidth-8pt\relax}>{\raggedright\arraybackslash}p{\dimexpr0.185\linewidth-8pt\relax}>{\raggedright\arraybackslash}p{\dimexpr0.185\linewidth-8pt\relax}}
\caption{Annotation efficiency compared with remote-sensing baselines}\label{tab:learning_baselines}\\*
\multicolumn{5}{p{\dimexpr\linewidth-8pt\relax}}{Count-MAE for all five methods using the same training subsets and test scenes; lower is better. These results use the same subsets as Supplementary Table~\ref{tab:learning_dense}, with 100, 400, 1,600 and all 4,627 training scenes. SIREN keeps its encoder and entity extraction heads fixed and trains damage and destruction classifiers on each subset. Each baseline is fine-tuned separately for each query using counts from the same subset.}\\*[5pt]
\multicolumn{5}{l}{\textbf{Damaged buildings}}\\*[4pt]
\toprule
Method & 100 scenes & 400 scenes & 1,600 scenes & All 4,627 \\*
\midrule
\endfirsthead
\multicolumn{5}{l}{Supplementary Table~\thetable{} (continued)}\\
\toprule
Method & 100 scenes & 400 scenes & 1,600 scenes & All 4,627 \\
\midrule
\endhead
\endfoot
\endlastfoot
SIREN & \textbf{6.116} & \textbf{3.955} & \textbf{3.150} & \textbf{2.895} \\*
SatlasPretrain & 8.095 & 4.708 & 3.564 & 3.145 \\*
RemoteCLIP & 7.583 & 5.752 & 4.102 & 3.357 \\*
SkyCLIP & 7.144 & 5.860 & 4.168 & 3.335 \\*
DINOv3 & 7.749 & 5.882 & 4.193 & 3.812 \\*
\bottomrule
\addlinespace[6pt]
\multicolumn{5}{l}{\textbf{Destroyed buildings without a nearby intact building}}\\*[4pt]
\toprule
Method & 100 scenes & 400 scenes & 1,600 scenes & All 4,627 \\*
\midrule
SIREN & \textbf{1.476} & \textbf{1.215} & \textbf{1.087} & \textbf{0.979} \\*
SatlasPretrain & 2.650 & 1.571 & 1.367 & 1.204 \\*
RemoteCLIP & 2.711 & 2.246 & 1.688 & 1.365 \\*
SkyCLIP & 2.689 & 2.179 & 1.614 & 1.333 \\*
DINOv3 & 2.816 & 2.328 & 1.827 & 1.494 \\*
\bottomrule
\addlinespace[6pt]
\end{longtable}
}

{\small
\setlength{\tabcolsep}{4pt}
\renewcommand{\arraystretch}{1.12}
\begin{longtable}{>{\raggedright\arraybackslash}p{\dimexpr0.28\linewidth-8pt\relax}>{\raggedright\arraybackslash}p{\dimexpr0.24\linewidth-8pt\relax}>{\raggedright\arraybackslash}p{\dimexpr0.24\linewidth-8pt\relax}>{\raggedright\arraybackslash}p{\dimexpr0.24\linewidth-8pt\relax}}
\caption{External disaster assessment}\label{tab:external_counting}\\*
\multicolumn{4}{p{\dimexpr\linewidth-8pt\relax}}{Count-MAE for destroyed buildings against independent damage records, without event-specific training or calibration. All methods use the same eligible image pairs within each event; lower is better. Reference building locations are not provided to the models. Error reductions are relative to the strongest baseline for each event.}\\*[5pt]
\toprule
Method & Hurricane Ian (2022; n=101) & Eaton fire (2025; n=131) & Palisades fire (2025; n=167) \\*
\midrule
\endfirsthead
\multicolumn{4}{l}{Supplementary Table~\thetable{} (continued)}\\
\toprule
Method & Hurricane Ian (2022; n=101) & Eaton fire (2025; n=131) & Palisades fire (2025; n=167) \\
\midrule
\endhead
\endfoot
\endlastfoot
SIREN & \textbf{3.218} & \textbf{8.015} & \textbf{1.677} \\*
SatlasPretrain & 5.723 & 12.595 & 2.054 \\*
RemoteCLIP & 6.881 & 14.412 & 1.910 \\*
SkyCLIP & 7.168 & 13.107 & 1.820 \\*
DINOv3 & 8.188 & 16.908 & 2.401 \\*
SIREN error reduction (\%) & 43.8 & 36.4 & 7.9 \\*
\bottomrule
\addlinespace[6pt]
\end{longtable}
}

\bigskip
{\small
\setlength{\tabcolsep}{4pt}
\renewcommand{\arraystretch}{1.12}
\begin{longtable}{>{\raggedright\arraybackslash}p{\dimexpr0.16\linewidth-8pt\relax}>{\raggedright\arraybackslash}p{\dimexpr0.13\linewidth-8pt\relax}>{\raggedright\arraybackslash}p{\dimexpr0.13\linewidth-8pt\relax}>{\raggedright\arraybackslash}p{\dimexpr0.37\linewidth-8pt\relax}>{\raggedright\arraybackslash}p{\dimexpr0.21\linewidth-8pt\relax}}
\caption{Visual classifier inputs, supervision and decision thresholds}\label{tab:neural_concepts}\\*
\multicolumn{5}{p{\dimexpr\linewidth-8pt\relax}}{Pre and post refer to imagery before and after the disaster. Each classifier predicts whether the listed property applies to an entity.}\\*[5pt]
\toprule
Visual property & Entity & Images & Label source & Decision threshold \\*
\midrule
\endfirsthead
\multicolumn{5}{l}{Supplementary Table~\thetable{} (continued)}\\
\toprule
Visual property & Entity & Images & Label source & Decision threshold \\
\midrule
\endhead
\endfoot
\endlastfoot
Damaged & Building & Pre and post & DisasterM3 damaged and destroyed labels & 0.55 \\*
Destroyed & Building & Pre and post & DisasterM3 destroyed labels & 0.60 \\*
Blocked & Road segment & Pre and post & DisasterM3 blocked-road labels & 0.83 (blocked-road counting); 0.45 (road access and qualitative examples) \\*
Residential & Building & Pre only & NSI; OpenStreetMap; FEMA USA Structures & 0.45 \\*
Surrounded by vegetation & Building & Pre only & GPT-5.4 labels, all validated by human review & 0.52 \\*
\bottomrule
\addlinespace[6pt]
\multicolumn{5}{p{\dimexpr\linewidth-8pt\relax}}{Thresholds for counting damaged buildings, destroyed buildings and blocked roads minimize validation \mbox{Count-MAE}. Residential and vegetation thresholds maximize validation macro-F1. The road-access query and the qualitative examples use the road classification threshold selected by maximizing macro-F1 on annotated validation road segments; roads below this threshold are classified as clear. The two blocked-road thresholds are applied to the same visual classifier. Annotation-efficiency experiments select damage thresholds separately, as described in Supplementary Section~\ref{supp:learning}.}\\*[5pt]
\end{longtable}
}

\needspace{0.62\textheight}
\bigskip
{\small
\setlength{\tabcolsep}{4pt}
\renewcommand{\arraystretch}{1.12}
\begin{longtable}{>{\raggedright\arraybackslash}p{\dimexpr0.34\linewidth-8pt\relax}>{\raggedright\arraybackslash}p{\dimexpr0.33\linewidth-8pt\relax}>{\raggedright\arraybackslash}p{\dimexpr0.33\linewidth-8pt\relax}}
\caption{Vision-language model evaluation settings}\label{tab:api_settings}\\*
\multicolumn{3}{p{\dimexpr\linewidth-8pt\relax}}{Evaluation settings for the two frontier vision-language models. Both receive paired pre- and post-disaster images, the same query wording and explicit definitions of the relevant properties and geometric conditions, without task-specific training.}\\*[5pt]
\toprule
Setting & Claude Opus 5.5 & GPT-6.1 Sol \\*
\midrule
\endfirsthead
\multicolumn{3}{l}{Supplementary Table~\thetable{} (continued)}\\
\toprule
Setting & Claude Opus 5.5 & GPT-6.1 Sol \\
\midrule
\endhead
\endfoot
\endlastfoot
API model ID & claude-opus-5-5 & gpt-6.1-sol \\*
Reasoning & Adaptive thinking, high effort & Low effort, selected on validation data \\*
Output limit & 32,000 tokens & 32,000 tokens \\*
Response format & Structured JSON; the count field is used for evaluation & Structured JSON; the count field is used for evaluation \\*
Image input & Full-size images supplied by file reference & Original image input \\*
Completed requests / failures & 12,819 / 0 & 12,819 / 0 \\*
\bottomrule
\addlinespace[6pt]
\multicolumn{3}{p{\dimexpr\linewidth-8pt\relax}}{\textbf{Claude prompt comparison.}}\\*[5pt]
\multicolumn{3}{p{\dimexpr\linewidth-8pt\relax}}{We compared direct-count, entity-listing and three-example prompts on 60 validation scenes per query. Entity-listing prompts request the positions of qualifying entities before returning a count. Three-example prompts include labeled validation examples that are excluded from the evaluation sample.}\\*[5pt]
\multicolumn{3}{p{\dimexpr\linewidth-8pt\relax}}{A failed answer counts as zero. We aggregate performance using the geometric mean of the per-query Count-MAE ratios relative to direct counting, adding 0.01 to each error before taking the ratio. The direct-count prompt has the lowest aggregate error and is used for both models.}\\*[5pt]
\toprule
Prompting strategy & \multicolumn{2}{p{\dimexpr0.66\linewidth-8pt\relax}}{Aggregate validation error ratio $\downarrow$} \\*
\midrule
Direct count & \multicolumn{2}{c}{\textbf{1.0000}} \\*
Entity listing & \multicolumn{2}{c}{1.1847} \\*
Three examples & \multicolumn{2}{c}{1.0178} \\*
\bottomrule\addlinespace[6pt]
\multicolumn{3}{p{\dimexpr\linewidth-8pt\relax}}{\textbf{GPT reasoning comparison.}}\\*[5pt]
\multicolumn{3}{p{\dimexpr\linewidth-8pt\relax}}{We compared high, medium and low reasoning effort on 60 validation scenes for counting damaged buildings and for counting destroyed buildings. We used low effort for the full-test evaluation. The validation results are shown below.}\\*[5pt]
\toprule
GPT reasoning effort & Damaged Count-MAE & Destroyed Count-MAE \\*
\midrule
High & 6.667 & 4.583 \\*
Medium & 7.517 & 5.167 \\*
Low & 7.017 & 3.867 \\*
\bottomrule
\end{longtable}
}

\clearpage
\bibliography{supplement-bibliography}